\documentclass[peerreview]{IEEEtran}
\usepackage[utf8]{inputenc}

\usepackage{xcolor}
\usepackage{algorithm,algorithmic}

\usepackage{multirow}
\usepackage{subfig}

\newtheorem{definition}{Definition}
\newtheorem{example}{Example}
\usepackage{array}

\usepackage{enumitem} 
\usepackage{balance}

\usepackage{amsmath}
\usepackage{amssymb}

\usepackage{graphicx} 

\usepackage{cite} 
\usepackage{url} 
\usepackage[utf8]{inputenc} 
\usepackage{booktabs} 
\usepackage{graphicx}
\usepackage{authblk}

\begin{document}
\title{A Report on Leveraging Knowlegde Graphs for Interpretable Feature Generation}

\author[1]{Mohamed Bouadi}
\author[2]{Arta Alavi}
\author[3]{Salima Benbernou}
\author[3]{Mourad Ouziri}
\affil[1]{SAP Labs Paris \& Université Paris Cité, LIPADE}
\affil[2]{SAP Labs Paris}
\affil[3]{Université Paris Cité, LIPADE}

\maketitle
\tableofcontents
\listoffigures
\listoftables

\IEEEpeerreviewmaketitle
\begin{abstract}
    The quality of Machine Learning (ML) models strongly depends on the input data, as such Feature Engineering (FE) is often required in ML. In addition, with the proliferation of ML-powered systems, especially in critical contexts, the need for interpretability and explainability becomes increasingly important. Since manual FE is time-consuming and requires case specific knowledge, we propose \textit{KRAFT}, an AutoFE framework that leverages a knowledge graph to guide the generation of interpretable features. Our hybrid AI approach combines a neural generator to transform raw features through a series of transformations and a knowledge-based reasoner to evaluate features interpretability using Description Logics (DL). The generator is trained through Deep Reinforcement Learning (DRL) to maximize the prediction accuracy and the interpretability of the generated features. Extensive experiments on real datasets demonstrate that \textit{KRAFT} significantly improves accuracy while ensuring a high level of interpretability.
\end{abstract}

\section{Introduction}
\begin{figure}[ht]
    \centering
    \includegraphics[width=1\linewidth, height=4cm]{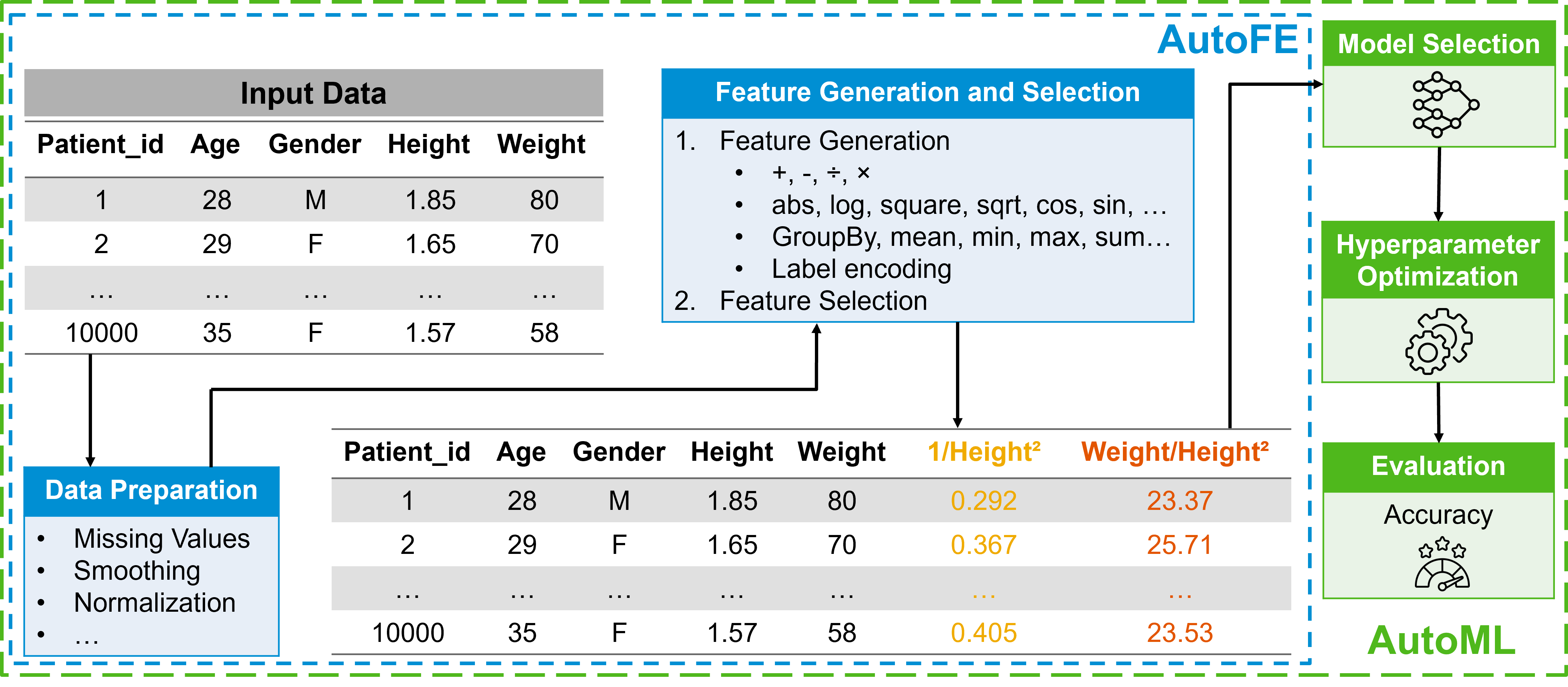}
    \caption{\centering Automated Machine Learning workflow.}
    \label{fig:example}
\end{figure}

In the last decade, ML has received an increasing interest. The reason is the resulting ability of organizations to automate the mundane and create business solutions to solve real world problems \cite{DBLP:journals/pvldb/00080LWZCDHWWZR21}. This success is often attributed to the experience of data scientists who leverage domain knowledge to extract useful patterns from data. This crucial yet tedious task, performed manually, is commonly referred to as Feature Engineering (FE). With limited human resources but ever-growing computing capabilities, AutoFE emerges as an important topic in AutoML, as shown in Figure \ref{fig:example}, as it may reduce data scientists workload significantly for quicker decision-making. Several approaches of AutoFE have been proposed \cite{khurana2018feature, chen2019neural, waring2020automated}. Nevertheless, these approaches have several drawbacks. For instance, they suffer from performance bottlenecks due to the large number of candidates. Additionally, they fall short in using external knowledge that is usually embedded in Ontologies and Knowledge Graphs \cite{DBLP:conf/semweb/DashBMG23,DBLP:conf/semweb/AsprinoC23}.

\begin{example} (Feature Generation).
Consider the problem of predicting heart disease of patients based on their attributes such as height, weight, age, amongst others. While these raw features provide valuable information, a more useful feature for this problem would be the \textit{Body Mass Index}, $BMI = \frac{weight}{height^2}$, which can be derived using two transformations – \textit{square} and \textit{division}, as shown in Figure \ref{fig:example} 
\end{example}

Explainable Artificial Intelligence (XAI) research has witnessed significant advancements, primarily focusing on the interpretability of ML models for domain experts \cite{jamshidi2023ecoshap,sivaprasad2023evaluation}. Recent studies have shown that the interpretability of ML models strongly depends on the interpretability of the input features \cite{zytek2022need}. However, existing AutoFE methods often struggle to discover easily understandable and interpretable features. This is due to the lack of research formalizing what makes a feature interpretable. Therefore, there is a real need for an approach that automates the generation of interpretable features for domain experts, to enhance their trust to the model's output \cite{zytek2022need}.

In recent years, Knowledge Ggraphs (KG) along with reasoning mechanisms have gained widespread adoption in a large number of domains \cite{DBLP:conf/semweb/IglesiasMolinaAIPC23,DBLP:conf/semweb/WuWLZWDHLR23}. KG and logical rules can provide context and semantic information about features and their valid combinations. This can be done through leveraging domain knowledge and linking data items with the semantic concepts and instances of the KG.

\textbf{Our work.} We introduce a knowledge-driven AutoFE approach that combines the use of a neural generator with a knowledge-based reasoner to maximize model accuracy while generating interpretable features for domain experts. We use DRL (\textit{the Generator}), more specifically, Deep Q-Network (DQN) \cite{chrysomallis2023deep,mnih2015human}, to iteratively generate features through interacting with the environment, and a knowledge-based selector (\textit{the Discriminator}) to select the subset of interpretable features using DL. To the best of our knowledge, our work is the first to address the trade-off between model accuracy and feature interpretability.

\textbf{Contributions.} We summarize our contributions as follows: 
\begin{enumerate}
    \item We define feature interpretability in ML, taking into account both features semantics and structure.
    
    \item We formalize FE as an optimization problem that maximizes the prediction accuracy while maintaining features interpretablity.

    \item We propose a knowledge-driven solution that takes advantage of the semantics embedded in the domain knowledge to automatically engineer interpretable features using a DRL agent.
 
    \item We conduct experiments on a large number of datasets to validate our solution in generating interpretable features.  
\end{enumerate}

The paper is organized as follows: in Section 2, we review the SOTA. Section 3 provides the data model and problem definition. In Section 4, we present our solution for interpretable AutoFE and, in section 5, we subject our approach to a set of experiments. Finally, Section 7 concludes the paper.

\section{Problem Definition}
In this work, we address the problem of engineering interpretable features using reasoning mechanisms based on the knowledge embedded in KG.

\subsection{Defining Interpretability}
Building a model to generate interpretable features requires a clear definition of interpretability. However, there is no consensus regarding the definition of interpretability in ML and how to evaluate it. Since a formal definition could be elusive, we have sought insights from the field of psychology.

\begin{definition}(Interpretability). In general, to interpret means "to explain the meaning" or "to present in understandable terms". In psychology \cite{lombrozo2006structure}, interperetability refers to the ability to understand and make sense of an observation. In the context of ML, the authors in \cite{doshi2017towards} defines interpretability as the "ability to explain or to present in understandable terms to a human".
\end{definition}

ML models need to learn from good data. Even simple and interpretable models, like regression, become difficult to understand with non-interpretable features. However, it is important to recognize that different users may have different requirements when it comes to feature interpretability. In our research, we are focusing on \textit{feature interpretability} for \textit{domain experts}, by capturing the connections between features semantics and domain knowledge.

\begin{definition}(Feature Interpretability). We define feature interpretability as the ability of domain experts to comprehend and connect the generated features with their domain knowledge. This implies mapping every new feature to relevant intensional and extensional knowledge within the domain of interest to gain deeper insights into the training data of the ML model. 
\label{def:interp}
\end{definition}

Consequently, interpretable features should be humanly-readable and refer to real-world entities that domain experts can understand and reason about. 

\subsection{Feature Engineering}
Consider a predictive problem on a tabular dataset $D=(X,Y)$ consisting of: 
\begin{enumerate}[label=(\roman*)]
    \item A set of features, $F^r \cup F^g$, where $F^r$ and $F^g$ are the raw features and the generated features respectively;
    \item An applicable ML algorithm, $L$, that accepts a training set and a validation set as input and returns the predicted labels, $y$, and a cross-validation performance measure $\mathcal{E}$;
    An interpretability function, $I_{KG} : Dom^F \rightarrow \{0,1\}$, where $Dom^F$ is the set of all possible features, that returns $1$ if the feature is interpretable according to the KG (i.e., it can be interpreted by a domain expert) and $0$ otherwise.

    \item A FE pipeline, $\mathcal{T} = \{t_1,..,t_m\}$, consisting of an ordered sequence of $m$ transformations; 
\end{enumerate}
     
Our problem is stated as finding a set of features, $F^*$, that maximizes the predictive accuracy of the given algorithm $L$ and ensures the requisite interpretability for large scale industrial tasks, i.e., the generated features must remain, as far as possible, close to the concepts known by domain experts.

\begin{equation}
    \label{eq:FE}
    \begin{split}
        \mathit{\mathcal{F}^* = \max_{ F \subseteq F^r \cup F^g}\mathcal{E}(L(F,Y))}\\
        \noindent s.t.\\
        \mathit{\prod_{f_i \in \mathcal{F}^*} \text{ } \mathcal{I}_{KG}(f_i) = 1},
       ~ \mathcal{I}_{KG}(f_i) \in \{0,1\}, \forall{f_i \in \mathcal{F}^*}\\
    \end{split}
\end{equation}

It should be pointed that, in real world, analysis tools are generic and are not dedicated to a specific business. Therefore, they must be able to exploit different data models, and sometimes cross multiple domains to address complex business issues, which means that FE may be conducted by domain experts who do not necessarily have skills in data science. Thus, in our work, we focused on developing a scalable solution to automate FE and more importantly to ensure the interpretability of the generated features, i.e., our solution is more likely to obtain features with statistical significance and remain interpretable by domain-experts.

\section{Proposed Approach}
To tackle the aforementioned problem, we propose \textit{KRAFT}. We first present an overview, then we elaborate on the components.

\begin{figure*}[ht!]
    \centering
    \includegraphics[width=1\linewidth,height=5.5cm]{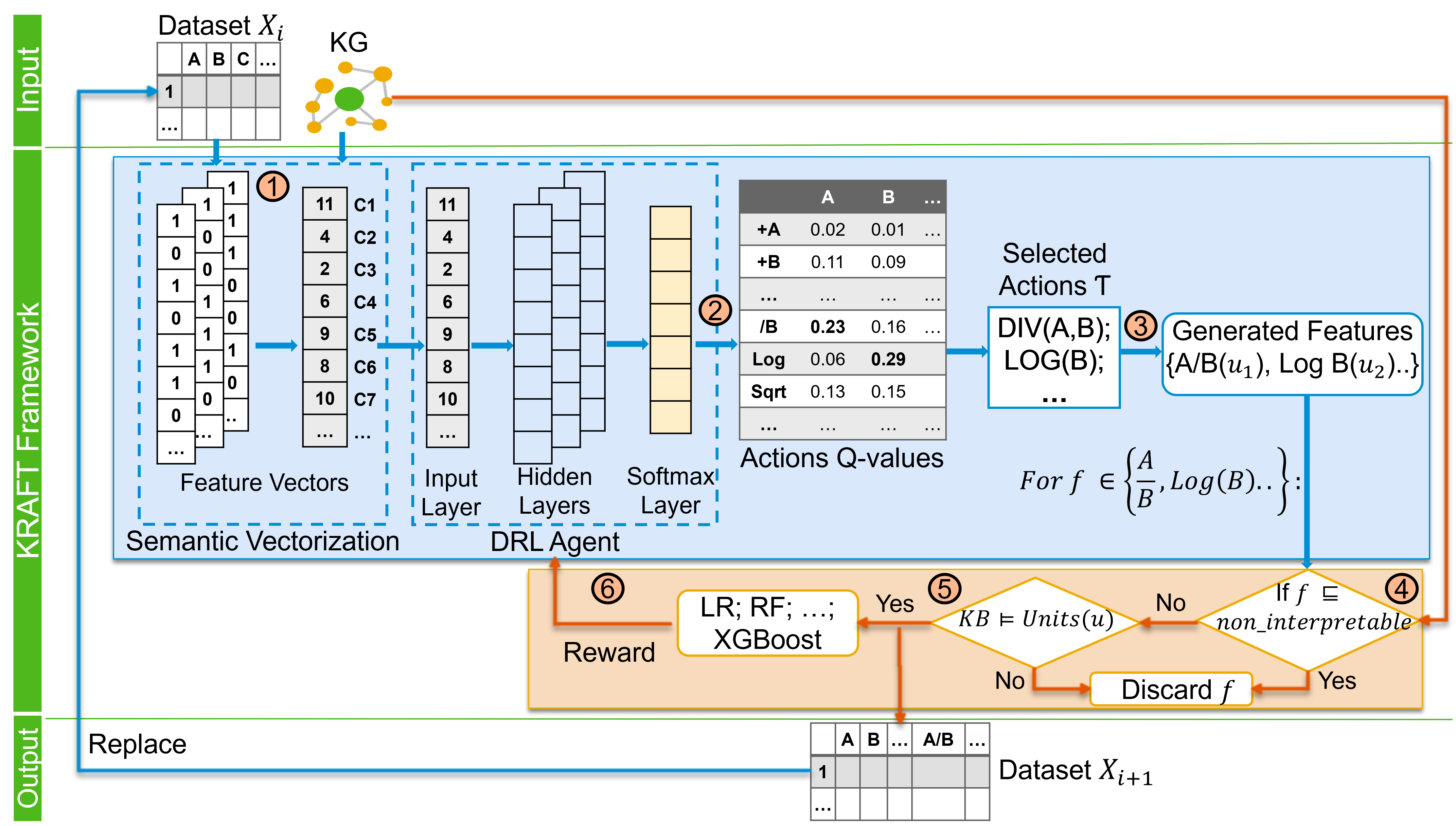}
    \caption{\centering An overview of \textit{KRAFT} architecture.}
    \label{fig:Archi}
\end{figure*}

\subsection{An Overview of \textit{KRAFT}}
As depicted in Figure \ref{fig:Archi}, \textit{KRAFT} consists of two parts: a \textit{Generator} (Blue) and a \textit{Knowledge-based Discriminator} (Orange). The former is a DRL agent that learns to generate interpretable features. We first start by performing a semantic vectorization to represent the input data with a feature vector based on the KG semantics (Figure \ref{fig:Archi}.1). Then, we use a DQN to estimate the transformations probabilities based on historical data (Figure \ref{fig:Archi}.2). The transformations with higher probability are used to generate new features (Figure \ref{fig:Archi}.3). The Knowledge-based Discriminator considers every generated feature as a DL concept and performs logical reasoning on the KG to select a subset of interpretable features from the generated ones (Figures \ref{fig:Archi}.4 and \ref{fig:Archi}.5). Then it rewards the generator for producing meaningful features. The reward is calculated by training an ML model on the new dataset (Figure \ref{fig:Archi}.6). This process continues iteratively until convergence.

\textit{KRAFT} can be seen as an enhancement of the Generative Adversarial Network (GAN) \cite{creswell2018generative}, which consists of two neural networks, a generator, that learns to generate plausible data, and a discriminator, that learns to distinguish between the generator's fake data and real data. Because it juggles two kinds of training, GAN frequently fails to converge \cite{wang2021generative}. However, \textit{KRAFT} uses a logical reasoner to distinguish interpretable features instead of training a neural network.

\subsection{Generator}
The goal of the generator is to learn how to generate interpretable features without exhaustively exploring all the search space. In the beginning, the search space is given by the following equation:

\begin{equation}
    \label{eq:searchSpace}
    Dom_F = \bigcup_{i=1}^{p} \text{ }\Biggl\{ \Biggl\{ \bigcup_{1 \leq s_1 \leq ..\leq s_i \leq p} \{ (f_{s_1},..,f_{s_i})\} \Biggl\} \times 
 \text{ } T_i \Biggl \},
\end{equation}

where $p$ is the number of features and $T_i \subseteq \mathcal{T}$ is the set of $i$-ary transformations.

The number of elements of this search space is:

\begin{equation}
    \label{eq:searchSpaceSize}
    |Dom_F| = \sum_{i=1}^{p} (A_i^p \times |T_i|),
\end{equation}

where $A_i^p$ is the $i$-permutation of $p$ features.

This space grows exponentially even with a limited number of transformations, hence an exhaustive search is not feasible. Therefore, in our work, we used DQN to learn an effective policy to iteratively generate new features.

\subsubsection{Semantic Vectorization}
\textit{KRAFT} addresses the limitations of traditional feature-based representations in neural networks, as they often lack rich semantics and domain knowledge. Instead of relying on numerical labels \cite{huang2022automatic,khurana2018feature,chen2019neural}, we propose a semantic vectorization to represent the input features with a feature vector based on the semantics embedded in our KG depicted in Figure \ref{fig:kg}. Basically, each feature $f$ in the dataset is represented with a feature vector $\Phi(f) \in R^n$. Each dimension of this vector corresponds to a concept within the KG. The dimensions relating to $f$ are set to $1$; i.e. its super-classes and unit of measurement, while the other dimensions are set to $0$. Since the number of features is different at each iteration and from a dataset to an other, a final feature vector for all input features, $\Phi(F)$, is obtained by summing the feature vectors: $\Phi(F) = \sum_{f \in F}\Phi(f)$, as shown in Figure \ref{fig:Archi}.1 . 

\subsubsection{Feature Generation}
We consider FE on a dataset $D$ as a Markovian Decision Process (MDP), which provides a mathematical framework for our decision problem. The goal of our agent is to find the best \textit{policy}, which is a mapping from \textit{states} to \textit{actions}, to maximize a cumulative \textit{reward}.

\textbf{State.} A state, $s_i \in S$, is the feature vector, $\Phi(D_i)$, representing the current dataset, created using the semantic vectorization described above.

\textbf{Action.} The set of actions, $A$, corresponds to the set of transformations $\mathcal{T}$. We distinguish different types of transformations: (i) Unary functions: such as \textit{logarithm} and \textit{square root} for numerical features and \textit{One-Hot-Encoding} for categorical features; (ii) Binary functions: like arithmetic operators ($+, -, \times, \div$) and logical operators ($\land, \lor,...$); (iii) Aggregation functions: using operators like \textit{Min}, \textit{Max}, \textit{Mean} and \textit{Sum}; (iv) Customized functions: such as \textit{Day}, \textit{Month}, \textit{Year} and \textit{Is\_Weekend}, to extract important information from dates features. When an action $a_i \in A$ is applied on state $s_i$, a new state $s_{i+1} = \{s_i, a_i(s_i)\}$ is generated, i.e., the new feature set is obtained by concatenating the features of the previous state $s_{i}$ with the new generated features $a_i(s_i)$, which can be seen as a Markov property.

\textbf{Reward.} The reward function consists of an evaluation metric, $\mathcal{E}$, and a pre-selected ML model, $L$. It is the average performance gained from the previous step on a k-fold cross validation. Formally, the reward can be written as:
 \begin{equation}
     \label{eq:reward}
     r_i = \mathcal{E}(L(F_{i+1})) - \mathcal{E}(L(F_{i}))
 \end{equation}

\textbf{Selecting an action.} To select an action, we used \textit{decaying $\epsilon$-greedy} \cite{wunder2010classes}. It consists of taking the action with the maximum reward with a probability of $1-\epsilon$, or a random action with a probability of $\epsilon$, while decreasing $\epsilon$ overtime. Thus, in the beginning we encourage the agent to better explore its environment. Once it learns, we decrease $\epsilon$ to allow the agent to exploit its knowledge.

\textbf{Policy.} The DRL agent is modeled by a policy network $\pi : S \rightarrow Q(A)$, where $Q$ is the expected cumulative reward estimated by the agent. The goal is to obtain the best sequence of $m$ transformations. We used two instances of a fully-connected multi-layer neural network to approximate the Q-function: (i) The main neural network, $Q(s,a;\theta_i)$, takes the current state and outputs an expected cumulative reward for each possible action; (ii) The target network, $\hat{Q}(s,a;\hat{\theta}_i)$, with delayed updates helps mitigating issues related to moving target distributions. A Q-learning update at iteration $i$ is defined as the loss function: 
\begin{equation}
    \label{loss}
    L_i(\theta_i) = \mathbb{E}_{MB}[(r+\gamma \max_{a'} \hat{Q}(s_{i+1},a';\hat{\theta_i})-Q(s_i,a_i;\theta_i))^2]
\end{equation}

We stored past experiences in a replay buffer and sampled them during training. This helps break the temporal correlation between consecutive samples, providing a more stable learning process.

Basically, at each step $i$ of the training, and iteratively until convergence, the agent receives the feature vector representing the current dataset, $\Phi(D_i)$, and feed it to the neural network. This latter calculates an intermediate reward for each possible action and selects the one that maximizes the long term reward. The selected transformation is then used to generate new features.

\subsection{Knowledge-based discriminator}
The discriminator's goal is to discard non-interpretable features from the set of the generated ones and then reward the agent accordingly.

\subsubsection{Knowledge representation}
\begin{figure}[ht]
    \centering
    \includegraphics[width=1\linewidth]{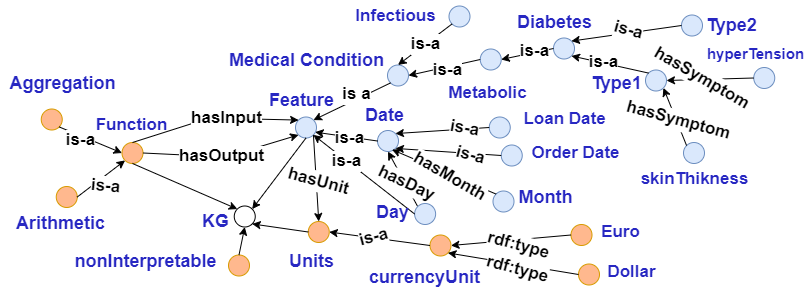}
    \caption{\centering A sample of the KG.}
    \label{fig:kg}
\end{figure}

The KG, displayed in Figure \ref{fig:kg}, contains two types of knowledge: (i) Domain-agnostic knowledge which includes knowledge about \textit{units of measurement} and \textit{quantities} from various domains (e.g. Physics, Geometry, the International System of Units), and semantics about arithmetic and aggregation transformations; (ii) Domain-specific knowledge that covers concepts from various business domains, such as Healthcare, Banking, Retail, E-commerce, Finance, and others. This part of the KG can be easily expanded over time to cover additional application areas using KG integration tools. The concepts of the KG are defined as DL concepts.   

\vspace{9pt}

In addition, we used SWRL (Semantic Web Rule Language) to define specific rules to determine whether a feature is interpretable. We show an example below. For instance, the first rule states that adding two features with different units would result in a non-interpretable feature and that periodic inventory totals are not summable. Similar explanations could be given to the other rules. 
\begin{align*}
(1)\text{ ~ } & Feature(?x) \land Feature(?y) \land Feature(?z) \land Addition(?f) \\
& \land hasUnit(?x, ?u) \land hasUnit(?y, ?v) \land Different(?u, ?v)   \land \\
& hasInput(?f, ?x) \land hasInput(?f, ?y) \land hasOutput(?f, ?z) \\
&  \rightarrow nonInterpretable(?z)\\
(2)\text{ ~ }  & aggregationSum(?f) \land Stock(?x) \land hasInput(?f,?x) \land \\
& Feature(?z) \land hasOutput(?f,?z) \rightarrow nonInterpretable(?z)\\
(3)\text{ ~ }  & Addition(?f) \land Temperature(?x) \land hasInput(?f,?x) \land  \\
& Feature(?z) \land hasOutput(?f, ?z) \rightarrow nonInterpretable(?z) \\
\end{align*}

\subsubsection{Feature Selection}
The discriminator is a reasoning algorithm used to discard non-interpretable features. We used HermiT \cite{shearer2008hermit}, a reasoner for the DL syntax, to infer new knowledge based on the logical relationships of the KG. To decide if a feature $f \in F$ with unit $u$ is interpretable, the discriminator considers $f$ as a DL concept and performs a subsumption reasoning. First, it checks if $f$ can be subsumed from the class \textit{non-interpretable}, i.e., $KB \models f \sqsubset$ \textit{non-interpretable}) (Figure\ref{fig:Archi}.4). In this case, $f$ would be removed. However, if we do not have enough information about the feature, the discriminator uses the knowledge about units and quantities that are stored as instances of the class \textit{Units}. For this, the model performs \textit{instance checking} ($KB \models Units(u)$) as shown in Figure \ref{fig:Archi}.5. If the unit is unknown, $f$ would be considered non-interpretable. In this way, the discriminator discards all non-interpretable features. Then a ML model is trained and evaluated on the remaining features and a reward is calculated as described in the previous section. 

Nevertheless, if $f$ is not covered by the KG and thus it is unknown to be interpretable or not, it will not be cut out from the features set. Similarly, if the KG is not available, \textit{KRAFT} will generate features with the sole aim of maximizing performance without considering feature interpretability. Consequently, the less knowledge we have about the dataset's domain, the fewer features we remove. This assumption helps avoid poor performance if the dataset is not properly covered by the KG. In real-world scenarios, a domain expert using \textit{KRAFT} could first check whether its KG covers the dataset. This could be done using coverage metrics \cite{jagvaral2020path} to evaluate the degree to which the KG covers the domain of interest. If the KG does not cover the domain, then it can easily be completed using an integration framework without affecting our model.

\section{Analysis and Interpretation}

\section{Experiments}

In our experiments we aim to address the following key research questions: \textbf{Q1:} How effective is our approach compared to SOTA? \textbf{Q2:} How does \textit{KRAFT} perform with different models? \textbf{Q3:} How good is the interpretability of the features generated by \textit{KRAFT}? \textbf{Q4:} How much efficient is \textit{KRAFT} compared to SOTA methods?

\subsection{Experimental Settings}

\underline{\textbf{Datasets.}} We used several public datasets that covered various data characteristics from various domains, e.g., the dataset size varying from the magnitude of $10^3$ to $10^7$. All the datasets are available on Kaggle\footnote{https://www.kaggle.com/}, UCIrvine\footnote{https://archive.ics.uci.edu/} and OpenML\footnote{https://www.openml.org/} databases. Table \ref{tab:ds} shows the statistics of a sample of the benchmark datasets. 

\begin{table}[!ht]
    \centering
    \caption{Statistics of the benchmark datasets.}
    \label{tab:ds}
    \resizebox{1\linewidth}{!}{
    \begin{tabular}{ l c c c c} 
        \hline
        Dataset & Source & Task & \#Rows & \#Features \\
        \hline
        Diabetes & Kaggle & Class. & 768 & 8 \\
        

        German Credit & UCIrvine & Class. & 1001 & 24 \\
        


        Wine Quality White & UCIrvine & Class. & 4900 & 12 \\ 

        Home Credit Default Risk & UCIrvine & Class. & 30000 & 25 \\
        
        Amazon Employee & Kaggle & Class. & 32769  & 9 \\ 
        
        Higgs Boson & UCIrvine  & Class. & 50000 & 28 \\

        Medical appointment & Kaggle & Class. & 110527 & 13 \\
        \hline

        
        Openml\_637 & OpenML & Reg. & 1000 & 25 \\

        Openml\_618 & OpenML & Reg. & 1000 & 50 \\

        Airfoil & UCIrvine & Reg. & 1503 & 5 \\

        Bikeshare DC & Kaggle & Reg. & 10886 & 11 \\

        NYC Taxi Ride & Kaggle & Reg. & 2792376 &14 \\
        \hline

\end{tabular}
}
\end{table}

\noindent \underline{\textbf{Baselines}.} We compared \textit{KRAFT} with several methods. \textbf{Base} is the original dataset without FE. \textbf{Random} consists of randomly applying transformations to raw features to generate new ones.\textbf{Deep Feature Synthesis (DFS)} applies all transformations on all features then performs feature selection \cite{kanter2015deep}. \textbf{AutoFeat} \cite{horn2019autofeat} is a popular Python library. \textbf{RL-based model (RLM)} builds a transformation graph and employs Q-learning techniques \cite{khurana2018feature}. \textbf{LFE} \cite{nargesian2017learning} trains an MLP and recommends the most useful transformations but only for classification tasks. \textbf{mCAFE} \cite{huang2022automatic} is an RL method based on monte carlo tree search. \textbf{NFS} \cite{chen2019neural} is a neural search based approach that uses several RNN-based controllers to generate transformation sequences for raw feature. \textbf{SAFE} \cite{shi2020safe} is an AutoFE approach designed for industrial tasks that uses a set of heuristics to generate features and a selection process to choose the most effective ones. \textbf{DIFER} \cite{zhu2022difer} is a feature evolution framework based on an encoder-predictor-decoder architecture.

 \noindent \underline{\textbf{Evaluation Metrics}.}
For effectiveness, we use different evaluation metrics for different tasks. We use the metric \textit{1 - rae} (relative absolute error) \cite{shcherbakov2013survey} for regression (Reg.) problems:
\begin{equation}
    \label{eq:metrics}
    1-rae = 1 - \dfrac{\sum{|\hat{y}-y|}}{\sum{|\bar{y}-y|}}
\end{equation}
where $y$ is the target, $\hat{y}$ is the model prediction and $\bar{y}$ is the mean of $y$. For classification (Class.) tasks, following \cite{khurana2018feature}, we use \textit{F1-score} (the harmonic mean of precision and recall).

\subsection{Effectiveness of \textit{KRAFT} (Q1 \& Q2 )}
\begin{table*}[htbp]
    \centering
    \caption{Comparing the effectiveness of \textit{KRAFT} with the baselines on benchmark datasets.}
    \label{tab:all}
    \resizebox{1\linewidth}{!}{
    \begin{tabular}{c c c c c c c c c c c|c c c c c c c c c c c}
        \hline
        Dataset & ALG & Base & Random & DFS & AutoFeat & RLM$^*$ & NFS & DIFER & mCAFE & KRAFT &Dataset & ALG & Base & Random & DFS & AutoFeat & RLM$^*$ & NFS & DIFER & mCAFE & KRAFT\\
        \hline
        
        \multirow{5}{*}{\textbf{Diabetes}} & RF & 0.740  & 0.670 & 0.737 & 0.767 & - & 0.786 & 0.798 & 0.813 & \textbf{0.832} & \multirow{5}{*}{\textbf{Medical}} & RF & 0.491 &  0.484 &  0.499 & 0.790 & - & 0.650 & 0.779 & 0.786 & \textbf{0.832} \\
        & DT & 0.732 & 0.598 & 0.732 & 0.741 & - & 0.770 & 0.776 & 0.798 & \textbf{0.813} & & DT & 0.478 & 0.480 & 0.487 & 0.780 & - & 0.634  & 0.729 & 0.753 & \textbf{0.824}\\
        & LR & 0.753 & 0.613 & 0.748 & 0.780 & - & 0.792 & 0.816 & 0.819 & \textbf{0.852} & & LR & 0.502 & 0.498 & 0.517 & 0.803 & - & 0.702 & 0.829 & 0.753 & \textbf{0.839} \\
        & SVM & 0.742 & 0.647 & 0.719 & 0.756 & - & 0.762 & 0.770 & 0.788 & \textbf{0.839} & \textbf{appointment} & SVM & 0.482 & 0.476 & 0.491 & 0.782 & - & 0.641 & 0.739 & 0.764 & \textbf{0.830} \\
        & XGB & 0.755 & 0.732 & 0.750 & 0.788 & - & 0.792 & 0. 813 & 0.819 & \textbf{0.854} & & XGB & 0.503 &0.503  & 0.515 & 0.801 & - & 0.713 & 0.830 & 0.838 & \textbf{0.842}\\
        \hline

        \multirow{5}{*}{\textbf{German Credit}} & RF & 0.661 & 0.655 & 0.680 & 0.760 & 0.724 & 0.781 & 0.777 & 0.770 & \textbf{0.782} & \multirow{5}{*}{\textbf{Home Credit}} & RF & 0.797 & 0.766 & 0.802 & 0.806 & \textbf{0.831} & 0.799 & 0.810 & 0.801 & 0.830\\
        & DT & 0.651 & 0.650 & 0.672 & 0.720 & - & 0.761 & 0.752 & 0.765 & \textbf{0.775} & & DT & 0.782 & 0.761 & 0.789 &0.7.92 & - & 0.792 & 0.802 & 0.789 & \textbf{0.828} \\
        & LR & 0.670 & 0.657 & 0.692& 0.767 & - & 0.788 & 0.773 & 0.782 & \textbf{0.801} & & LR & 0.801 & 0.783 & 0.804 & 0.805 & - & 0.807 & 0.811 & 0.805 & \textbf{0.839} \\
        & SVM & 0.655 & 0.661 & 0.679 & 0.734 & - & 0.765 & 0.762 & 0.772 & \textbf{0.789} & \textbf{Default Risk}& SVM & 0.789 & 0.769& 0.793& 0.790& - & 0.801 & 0.806 & 0.803 & \textbf{0.827}\\
        & XGB & 0.690 & 0.666 & 0.695 & 0.770 & - &  0.790 & 0.792 & 0.790 & \textbf{0.806} & & XGB & 0.802 & 0.780  & 0.804 & 0.811 & - & 0.819 & 0.814 &0.809 & \textbf{0.842} \\
        \hline

        \multirow{5}{*}{\textbf{Wine Quality}} & RF & 0.494 & 0.482 & 0.488 & 0.502 & 0.722 & 0.516 & 0.515 & 0.502 &\textbf{0.732}&  \multirow{5}{*}{\textbf{Openml\_618}} & RF & 0.428 & 0.428 & 0.411 & 0.632 & 0.587 & 0.640 & 0.660 & \textbf{0.738} & \textbf{0.738}\\
        & DT & 0.492 & 0.473 & 0.475 &0.489 & - & 0.513 & 0.513 & 0.492 & \textbf{0.713 }& & DT & 0.425 & 0.419 & 0.408 & 0.630 & - &  0.629 & 0.645 & 0.725& \textbf{0.732} \\
        & LR & 0.501  & 0.489 & 0.490 &0.505 & - &  0.519& 0.517 & 0.509 & \textbf{0.727} & & LR &0.432 & 0.429&0.412 &0.645 & - & 0.642 & 0.666& 0.740& \textbf{0.743}\\
        \textbf{White}& SVM & 0.489 & 0.480 & 0.482 & 0.493 & - & 0.521 & 0.515& 0.499& \textbf{0.719} & & SVM & 0.423& 0.421& 0.405 & 0.635 & - & 0.632 & 0.643 & 0.720 & \textbf{0.733} \\
        & XGB & 0.503& 0.490& 0.493& 0.508 & - & 0.525&0.524 &0.511 & \textbf{0.735} & & XGB & 0.430 & 0.429 & 0.413 & 0.642 & - & 0.645 & 0.663 & 0.735 & \textbf{0.741}\\
        \hline

        \multirow{5}{*}{\textbf{Amazon}} & RF & 0.712 & 0.740 & 0.744 & 0.739 & 0.806 & \textbf{0.945} & 0.909 & 0.897 & 0.916 & \multirow{5}{*}{\textbf{Bikeshare DC}} & RF & 0.393 & 0.381 & 0.693 & 0.849 & 0.798 & 0.974 & \textbf{0.981} & 0.906 & 0.960\\
        & DT & 0.701 & 0.736 & 0.738 & 0.732 & - & \textbf{0.932} & 0.903 & 0.878 & 0.910 & & DT & 0.378 & 0.372 & 0.682 & 0.840 & - & \textbf{0.965} & 0.959 & 0.896 & 0.956 \\
        & LR & 0.715 & 0.744 & 0.750 &0.742 & - & \textbf{0.935}& 0.913 & 0.905 & 0.925 & & LR & 0.401 & 0.385 & 0.702 & 0.853 & - & 0.975 & \textbf{0.981} & 0.921 & \textbf{0.981}\\
        \textbf{Employee}& SVM & 0.710 & 0.742 &0.740 & 0.735& - & \textbf{0.936} & 0.910& 0.906 & 0.927& & SVM & 0.380& 0.381& 0.695& 0.845& - &0.978 & 0.978&0.922 &\textbf{0.980} \\
        & XGB &0.719 &0.748 &0.755 &0.745 & - & \textbf{0.945} & 0.915& 0.908& 0.936 & & XGB & 0.405& 0.389& 0.712& 0.855& - & 0.980& \textbf{0.982}& 0.932& 0.981\\
        \hline

        \multirow{5}{*}{\textbf{Higgs Boson}} & RF & 0.718 & 0.699 & 0.682 & 0.468 & 0.729 & 0.731 & 0.738 & \textbf{0.739} & 0.729 & \multirow{5}{*}{\textbf{NYC Taxi}} & RF &  0.425 & 0.434 & 0.505  & - & - & 0.551 & 0.501 & 0.446 & \textbf{0.573}\\
        & DT & 0.709& 0.684& 0.678& 0.460& - &0.725 &0.719 & \textbf{0.728}&0.725 & & DT & 0.419 & 0.429 & 0.489 & - & - & 0.539& 0.487& 0.426& \textbf{0.553}\\
        & LR & 0.722& 0.704& 0.700& 0.470& - &0.734 &\textbf{0.742} &\textbf{0.742} & 0.739& & LR & 0.430& 0.449 & 0.513 & 0.458 & - & 0.559 & 0.515 & 0.452 &\textbf{0.578} \\
        \textbf{Ride}& SVM &0.711 &0.705 &0.689 &0.465 & - &0.726 & 0.724& \textbf{0.732}& \textbf{0.732}& & SVM & 0.423 & 0.432 & 0.499 & - & - & 0.545 & 0.492 & 0.459& \textbf{0.562}\\
        & XGB &0.725 & 0.705& 0.706& 0.489& - & 0.735& 0.740& 0.735&\textbf{0.741} & & XGB & 0.432 & 0.459 & 0.515 & - & - & 0.562 & 0.523 & 0.486 & \textbf{0.588} \\
        \hline
    \end{tabular}
}
\end{table*}

\begin{table}[!ht]
  \centering
   \caption{Comparison between \textit{KRAFT}, \textit{LFE}, \textit{NFS} and \textit{DIFER}.}
   \label{tab:lfe}
  \resizebox{1\linewidth}{!}{
 \begin{tabular}{ l c c c c c} 
 \hline
  Dataset & Base & LEF & NFS & DIFER & KRAFT \\
 \hline
 Credit-a & 0.837 & 0.771 & 0.865 & 0.882 & \textbf{0.883}\\
 Feritility & 0.853 & 0.873 & 0.870 & \textbf{0.909} & 0.880\\
  Hepatitis & 0.786  & 0.831 & 0.877 & 0.883 & \textbf{0.887}\\
 Spam Base & \textbf{0.955} & 0.947 & \textbf{0.955} & 0.934 & \textbf{0.955}\\
 \hline
\end{tabular}
}
\end{table}

\begin{table}[!ht]
  \centering
   \caption{Comparing \textit{KRAFT} with \textit{SAFE} using AUC.}
   \label{tab:safe}
  \resizebox{1\linewidth}{!}{
 \begin{tabular}{ l c c c c} 
 \hline
  Dataset & Base & Random & SAFE & KRAFT \\
 \hline
Gina & 0.905 & 0.904 & 0.918 & \textbf{0.920}\\
 Bank &  0.665 & 0.675 & 0.682 & \textbf{0.711}\\
 Wind & 0.836 & 0.851 & \textbf{0.852} & 0.841\\
 Ailerons & 0.844 & 0.860 & 0.868 & \textbf{0.885}\\
 Vehicle & 0.855 & 0.862 & \textbf{0.865} & 0.864\\
 \hline 
\end{tabular}
}
\end{table}

We evaluate each baseline, when it was possible, on 5 SOTA ML models, which are: Random Forest (RF), Decision Tree (DT), Logistic/Linear Regression (LR), Support Vector Machines (SVM) and XGBoost (XGB). All parameters are set to default values in scikit-learn \cite{pedregosa2011scikit}. We performed the experiments several times and then averaged the results. We report in Table \ref{tab:all} the comparison of 10 datasets  (out of 26 due to lack of space). Since \textit{LEF} \cite{nargesian2017learning} can only deal with classification and there is no source code available for \textit{SAFE} \cite{shi2020safe}, we directly refer to the results reported in their papers. The comparison between \textit{KRAFT} and both \textit{LFE} and \textit{SAFE} is shown in Table \ref{tab:lfe} and Table \ref{tab:safe} respectively. 

Despite having an additional constraint of interpretability, our approach outperforms the baselines in most cases. Compared to raw data, the features generated by \textit{KRAFT} can improve the performance by an average of $18.53\%$. Compared with \textit{DIFER}, \textit{NFS}, \textit{mCAFE} and \textit{SAFE}, \textit{KRAFT} achieves an average improvement of $6.30\%$, $6.44\%$, $5.41\%$ and $2.58\%$ respectively. We show also in these experiments that, despite varying performance among different models, \textit{KRAFT} has a significant advantage over the baselines, emphasizing its model-agnostic nature.

\begin{figure}[ht]
    \centering
    \includegraphics[width=1\linewidth,height=3cm]{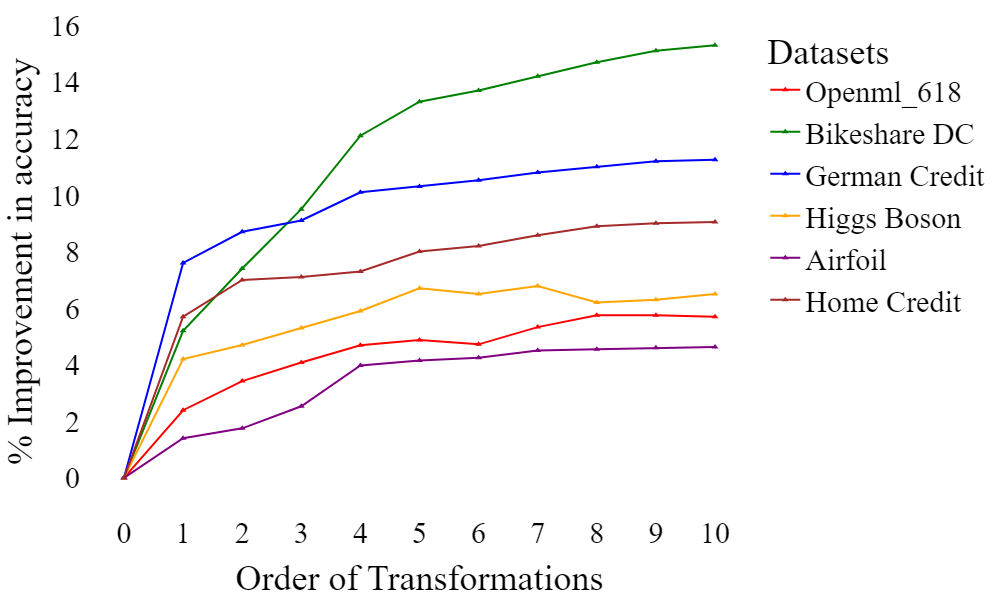}
    \caption{\centering Effect of high-order features on accuracy improvement.}
    \label{fig:orderperf}
\end{figure}

The performance of \textit{KRAFT} could be first explained by its use of semantic vectorization that leverages domain knowledge to generate features with richer information, which is the exact definition of FE. As explained before, the KG provides valuable insights about the semantics of features at different levels of abstraction. By feeding the semantic feature vectors into our learning algorithm, \textit{KRAFT} generates promising features, leading to better performance compared to the baselines. Moreover, \textit{KRAFT}'s ability to generate high-order features plays an important role in its performance. In fact, we show in Figure \ref{fig:orderperf} that as the maximum order of features increases, the accuracy of \textit{KRAFT} improves. This indicates the effectiveness of generating high-order features, as the composition of transformations is crucial for discovering complex relationships between features. 

\subsection{Interpretability Evaluation of \textit{KRAFT} (Q3)}

In our work, we focus on feature interpretability for domain experts, rather than the interpretability of the model itself. Our model generates features that are both readable, enabling users to understand their meaning, and interpretable, aligning with the concepts known and used by domain experts.

\begin{figure*}[!ht]
    \centering
    \captionsetup[subfigure]{labelformat=empty}
  \subfloat[Amazon Employee\label{amazonimp}]{%
  \includegraphics[width=0.32\linewidth, height=3cm]{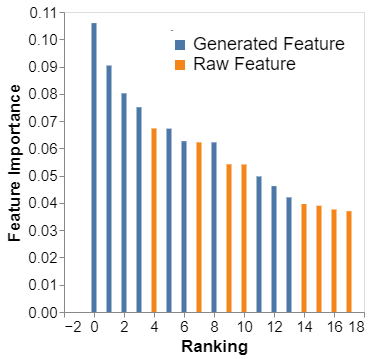}}
    \hfill
  \subfloat[Medical Appointment\label{medimp}]{%
  \includegraphics[width=0.32\linewidth, height=3cm]{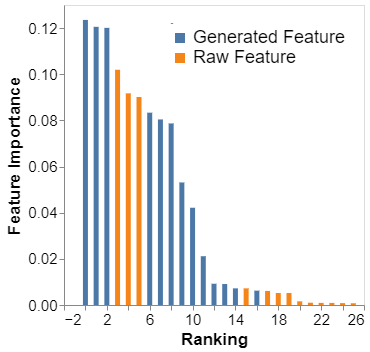}}
    \subfloat[NYC Taxi Ride\label{fnycimp}]{%
       \includegraphics[width=0.32\linewidth, height=3cm]{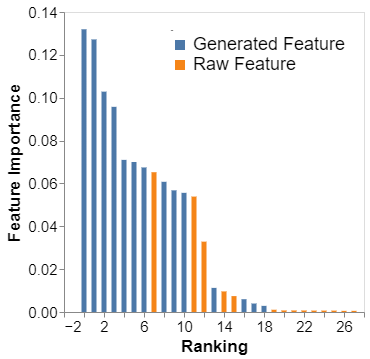}}\\
    \vspace{10pt}
  \subfloat[Home Credit\label{homeimp}]{%
        \includegraphics[width=0.32\linewidth,height=3cm]{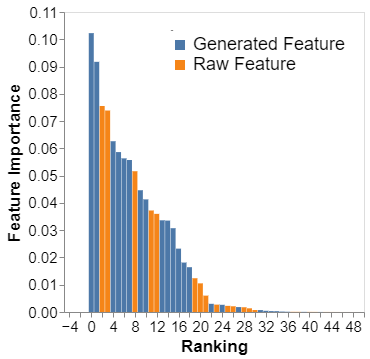}}
    \hfill
    \subfloat[Diabetes\label{diabimp}]{%
        \includegraphics[width=0.32\linewidth, height=3cm]{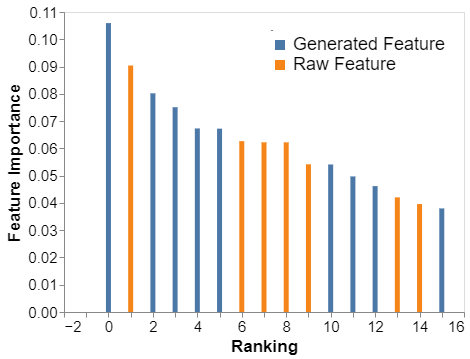}}
    \hfill
    \subfloat[Spam Base\label{spamimp}]{%
        \includegraphics[width=0.32\linewidth, height=3cm]{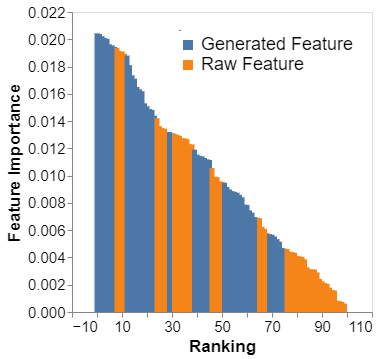}}\\
    \vspace{10pt}
    \subfloat[Higgs Boson\label{higgsimp}]{%
        \includegraphics[width=0.32\linewidth, height=3cm]{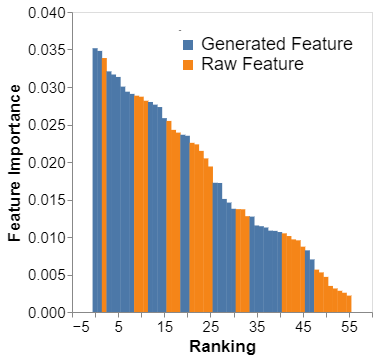}}
    \hfill
        \subfloat[Wind\label{windimp}]{%
        \includegraphics[width=0.32\linewidth, height=3cm]{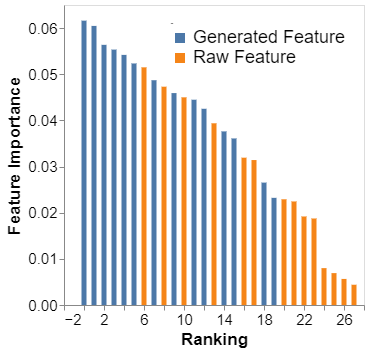}}
    \hfill
    \subfloat[Ailerons\label{ailerimp}]{%
        \includegraphics[width=0.32\linewidth, height=3cm]{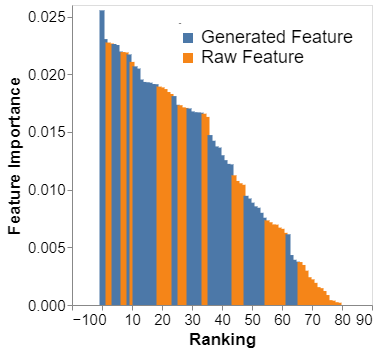}}
    \vspace{10pt}
  \caption{Comparing feature importance of raw features (in orange) and generated features with \textit{KRAFT} (in blue).}
  \label{fig:featuresImpranking} 
  \vspace{10pt}
\end{figure*}

\begin{figure*}[ht]
    \centering
    \captionsetup[subfigure]{labelformat=empty}
    
    \subfloat{\includegraphics[width=0.33\linewidth, height=4cm]{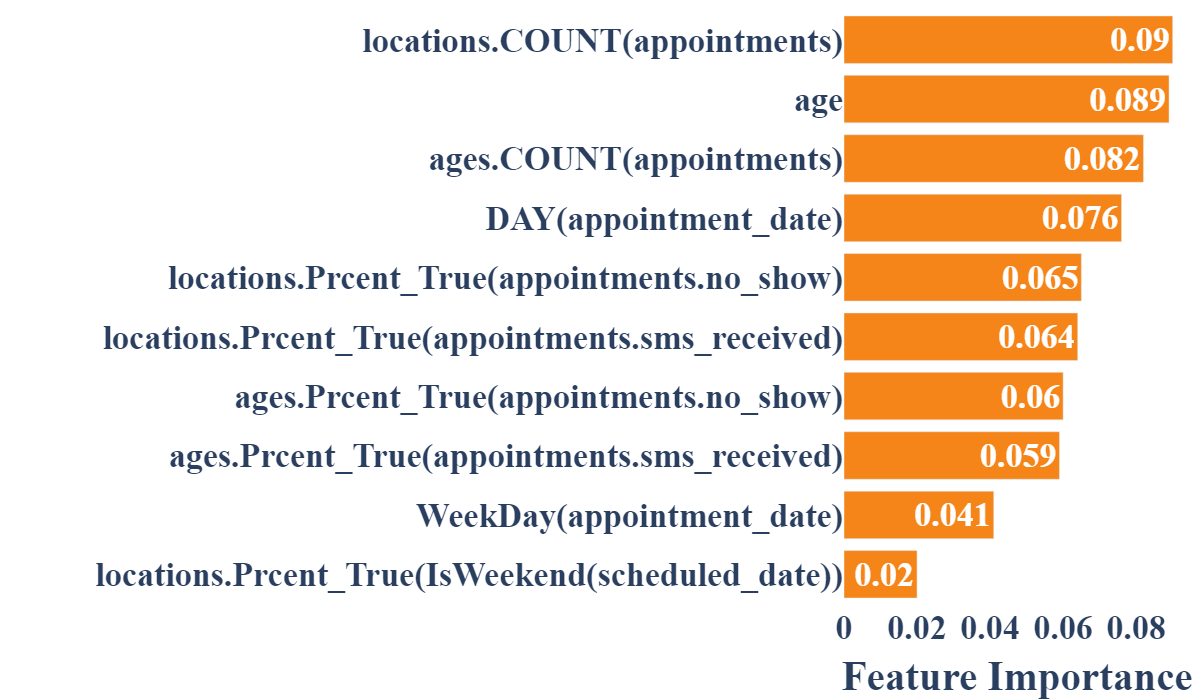}\label{fig:med_dsm}}
    \hfill
    \subfloat{\includegraphics[width=0.33\linewidth, height=4cm]{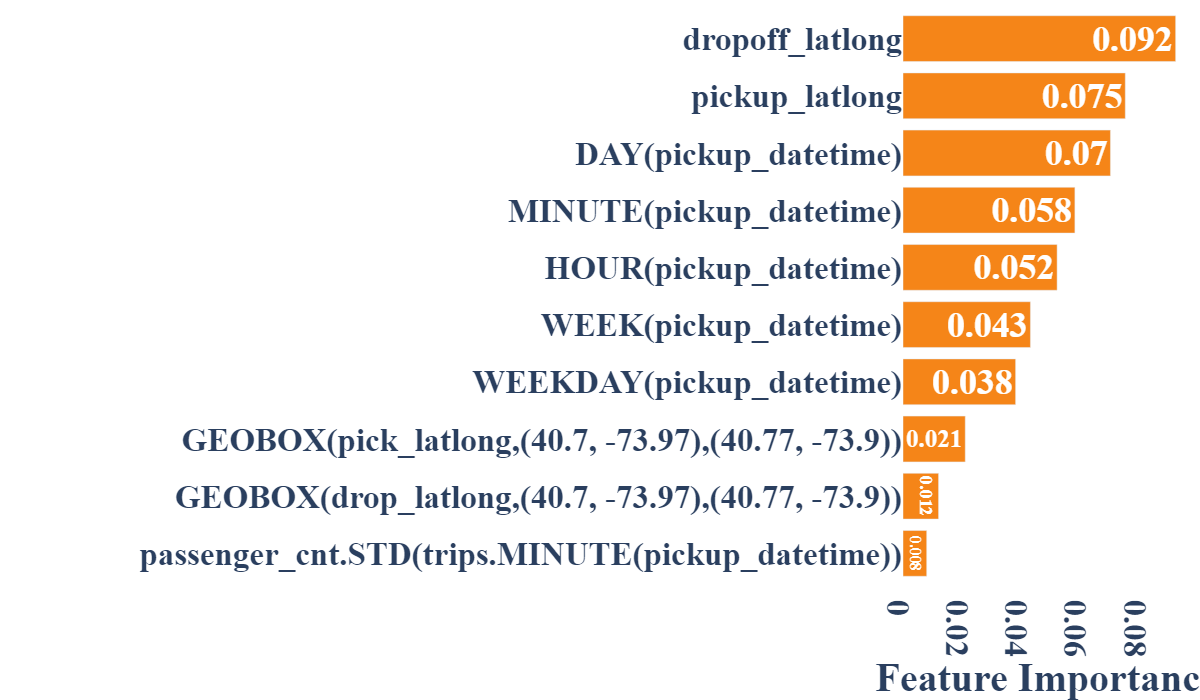}\label{fig:nyc_dsm}}
    \hfill
    \subfloat{\includegraphics[width=0.33\linewidth, height=4cm]{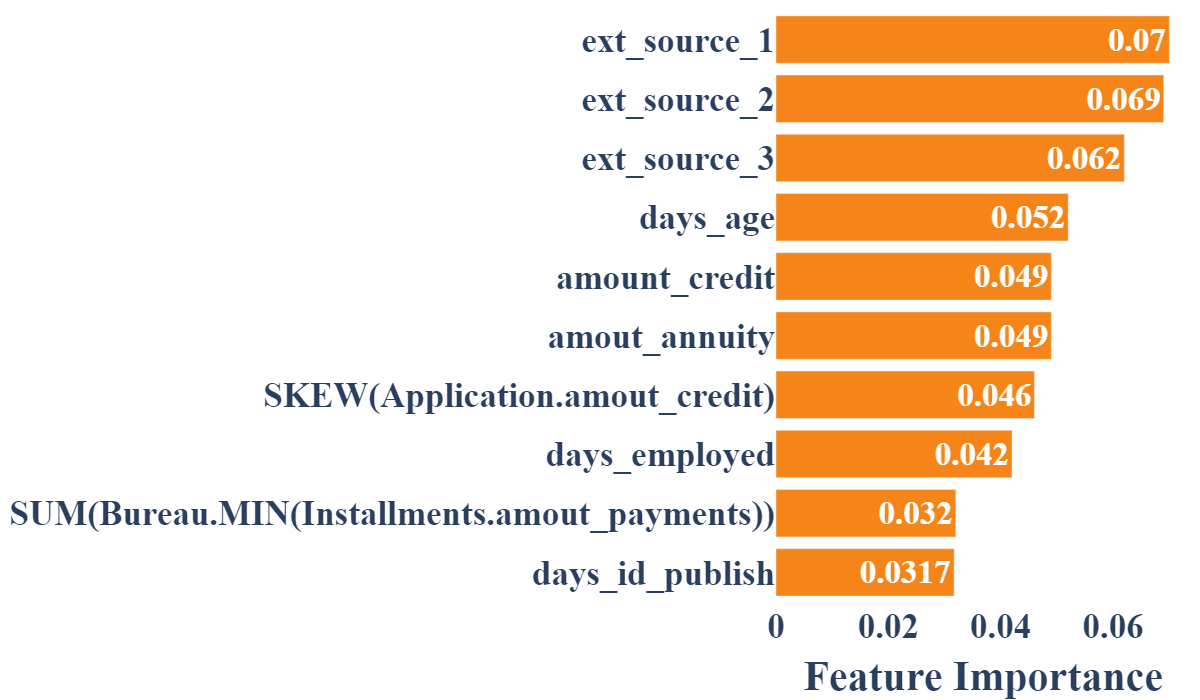}\label{fig:home_dsm}}
    \\
    \subfloat{\includegraphics[width=0.33\linewidth, height=4cm]{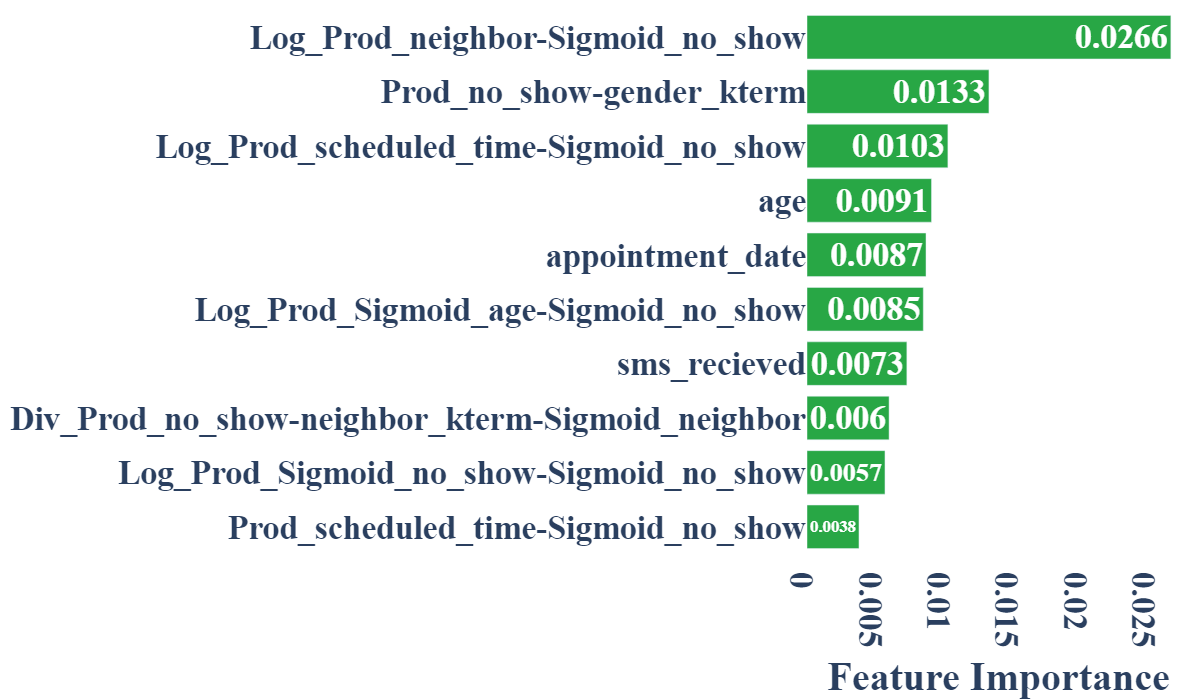}\label{fig:med_mcafe}}
    \hfill
    \subfloat{\includegraphics[width=0.33\linewidth, height=4cm]{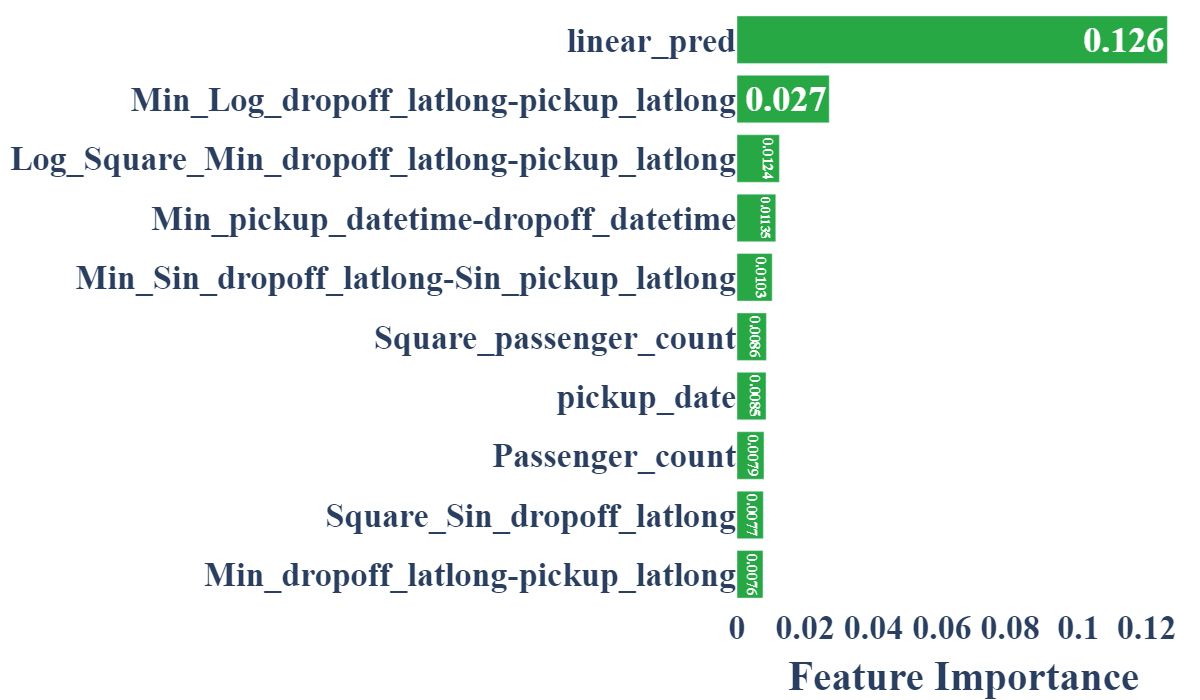}\label{fig:nyc_mcafe}}
    \hfill
    \subfloat{\includegraphics[width=0.33\linewidth, height=4cm]{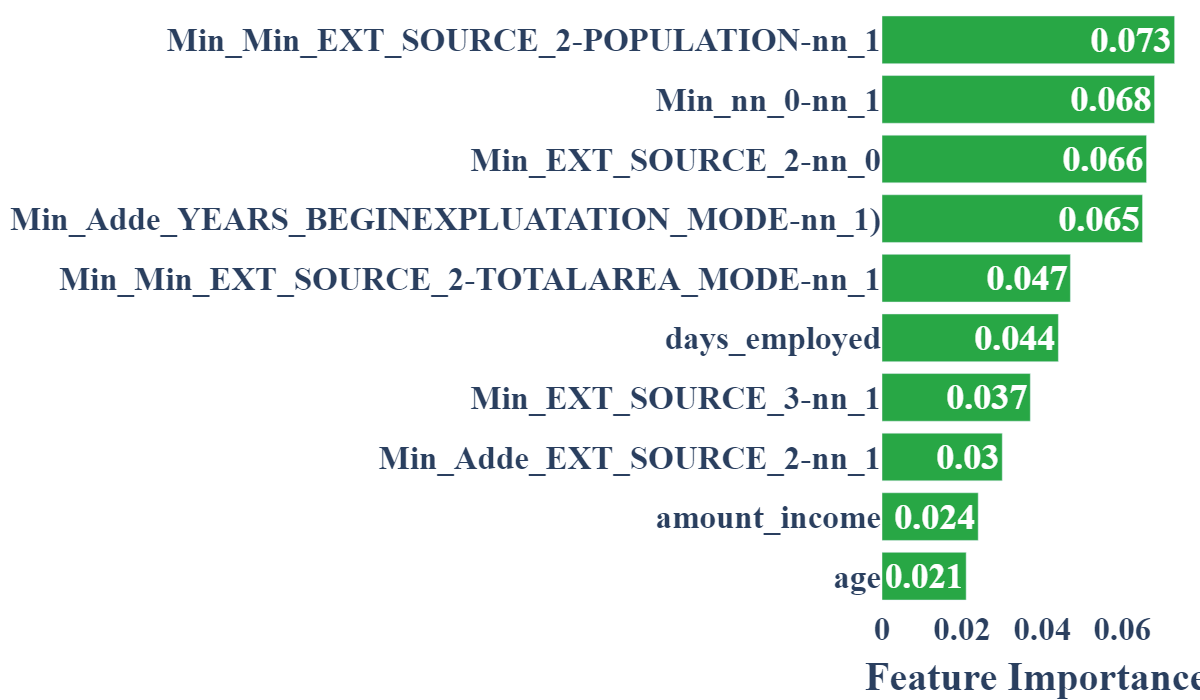}\label{fig:home_mcafe}}
    \\
    \subfloat[\textit{  } \textit{  }  \textit{  }  \textit{  } 
 \textit{  }  \textit{  }  \textit{  }  \textit{  } \textit{  } \textit{  } \textit{  } \textit{  } Medical Appointment]{\includegraphics[width=0.33\linewidth, height=4cm]{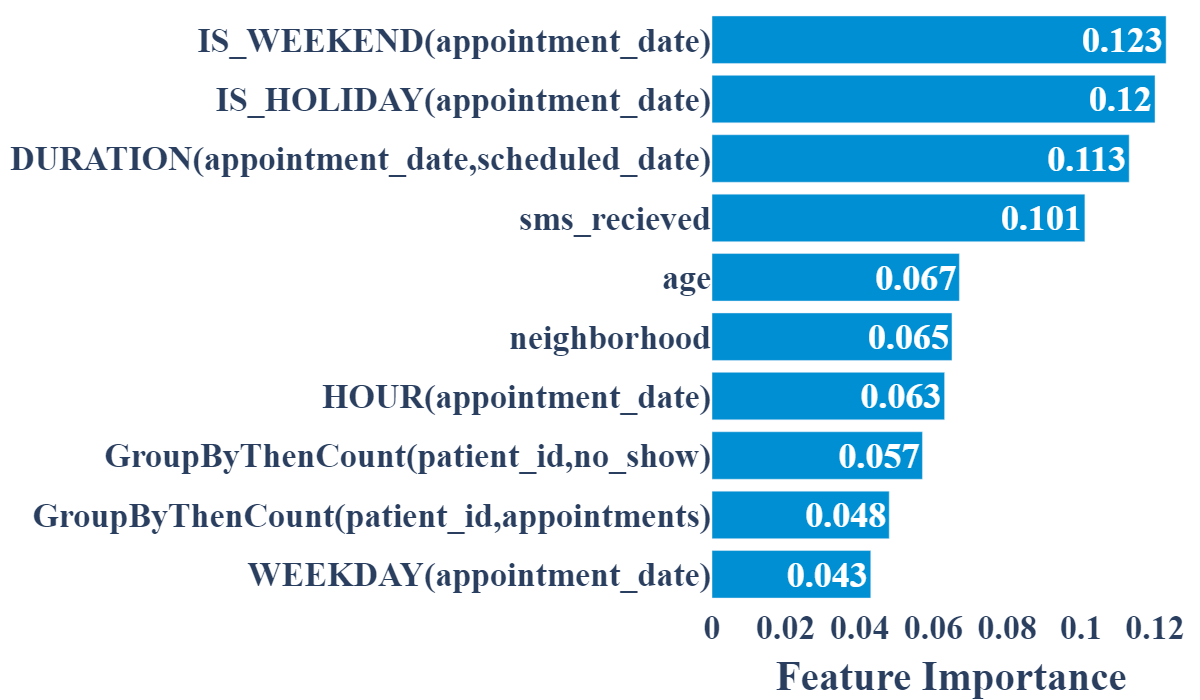}\label{fig:med_KRAFT}}
    \hfill
    \subfloat[\textit{  } \textit{  } \textit{  } \textit{  } \textit{  } \textit{  } \textit{  } \textit{  } NYC Taxi Ride Duration]{\includegraphics[width=0.33\linewidth, height=4cm]{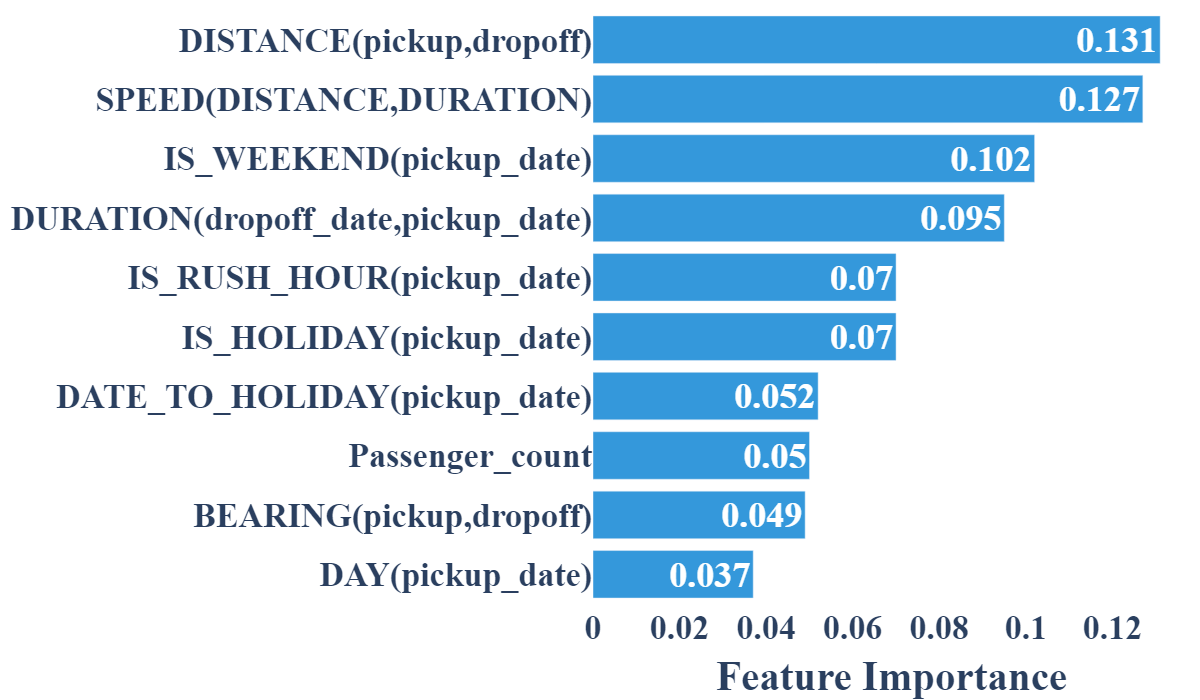}\label{fig:imp_nyc}}
    \hfill
    \subfloat[\textit{  } \textit{  } \textit{  } \textit{  } \textit{  } \textit{  } \textit{  } \textit{  } Home Credit Default Risk]{\includegraphics[width=0.33\linewidth, height=4cm]{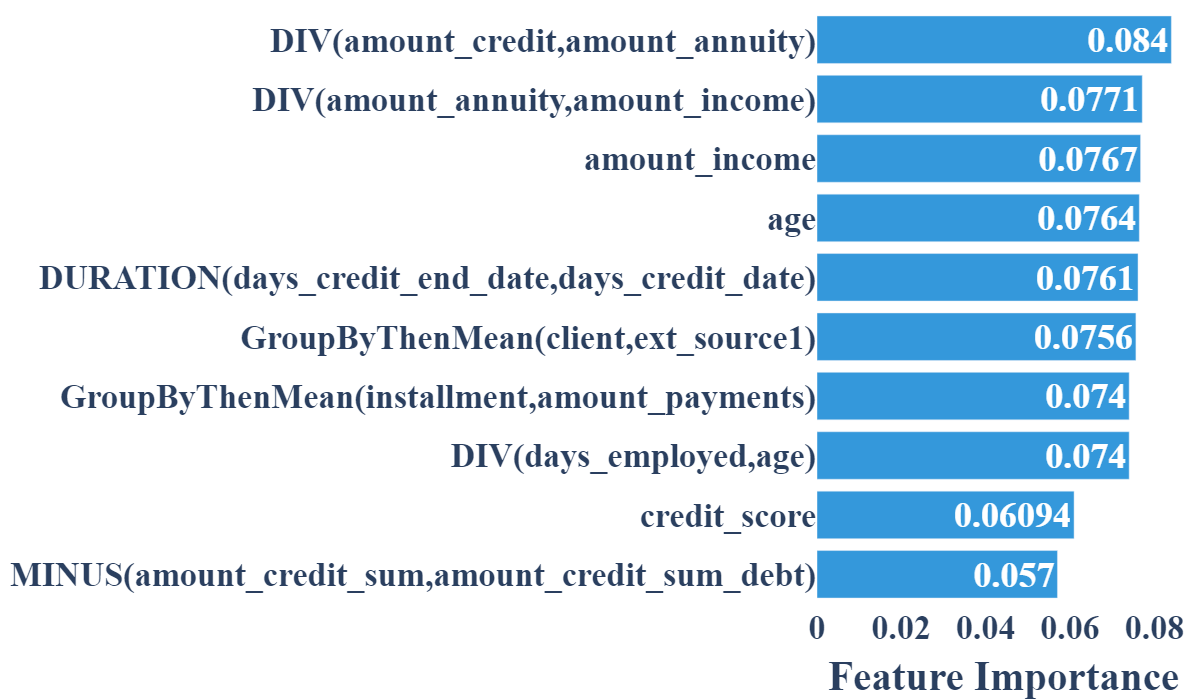}\label{fig:home_KRAFT}}

    \vspace{10pt}
    
    \caption{Comparing the top-10 important features between \textit{DSM} (in orange), \textit{mCAFE} (in green) and \textit{KRAFT} (in blue).}
    \label{fig:featuresImpo} 

    \vspace{10pt}
\end{figure*}

\textbf{Feature Importance.} In this experiment, we use RF to compare the importance of the generated features with raw features. We start by creating a new dataset combining the $n$ raw features with the top-ranked $n$ features generated by \textit{KRAFT}. We then use RF to score feature importance in the model's predictions. Figure \ref{fig:featuresImpranking} showcases the results for $4$ different datasets. Notably, the features generated by \textit{KRAFT} (shown in blue) are more important compared to raw features (shown in orange) across all datasets, thereby validating the effectiveness of \textit{KRAFT} in generating interpretable features.

\textbf{Model-agnostic Interpretability.} It refers to the ability to understand and interpret the decisions made by an ML model without relying on the specific details of its internal architecture. A common way to achieve this is by using the well-known approach SHAP (SHapley Additive exPlanations) \cite{lundberg2017unified} to quantify the contribution of input features to the model prediction, i.e., to attribute the predictions to specific parts of the input features. 

For a more in-depth analysis, we compare in Figure \ref{fig:featuresImpo} the top $10$ features generated by our model (in blue) with those generated by \textit{mCAFE} (in green) across three different datasets from different domains. It can be seen that, compared to the baseline, \textit{KRAFT} generates easily readable and interpretable features, aligning with the domain knowledge. For instance, in the \textit{NYC Taxi Ride} dataset, our model generated $9$ out of the top $10$ features, including crucial indicators like \textit{DISTANCE} and \textit{DURATION} between pickup and drop-off locations and times, directly correlating with ride duration. Moreover, it incorporates temporal aggregations such as \textit{IS\_RUSH\_HOUR} and \textit{DAYS\_TO\_HOLIDAY}, reflecting their significant impact on traffic conditions. However, features generated by the baseline remain challenging for domain experts to understand, and often they do not relate to the domain knowledge. In the \textit{Medical Appointment} dataset, \textit{KRAFT} generates insightful features such as \textit{DURATION} between appointment scheduling and the actual appointment, indicating a correlation with appointment no-shows. Additionally, it captures factors like \textit{IS\_WEEKEND} and \textit{IS\_HOLIDAY}. A thorough analysis of this dataset shows that patients have a higher tendency to miss their appointments on weekdays rather than on weekends or holidays.

\section{Conclusion}
Our work aims at bridging the gap between statistical ML and symbolic. Along these lines, we presented a hybrid approach to efficiently engineer interpretable features for domain experts from structured knowledge. The cornerstone of our approach are - a DRL-based exploration of the feature space, and a human-like reasoner over a KG using DL to remove non-interpretable features. Extensive experiments on large scale datasets are conducted, and detailed analysis is provided, which shows that the proposed method can provide competitive performance and guarantee the interpretability of the generated features. In this work, we modeled the interpretability problem as a binary decision problem and as a future work, we will go towards a flexible graduated framework of interpretability evaluation.

\bibliographystyle{ACM-Reference-Format}
\bibliography{main}

\end{document}